\documentclass[letterpaper, 10 pt, conference]{ieeeconf}
\IEEEoverridecommandlockouts                              

\usepackage[utf8]{inputenc} 
\usepackage[T1]{fontenc}    
\usepackage{hyperref}       
\usepackage{url}            
\usepackage{booktabs}       
\usepackage{amsfonts}       
\usepackage{nicefrac}       
\usepackage{microtype}      
\usepackage{xcolor}         
\usepackage{soul}           
\usepackage{multirow}
\usepackage{booktabs}
\usepackage{makecell}
\usepackage{amsmath}
\usepackage{cite}
\usepackage{soul}
\usepackage{pifont}

\usepackage{graphicx}
\usepackage[caption=false]{subfig}

\usepackage{algorithm}
\usepackage[noend]{algpseudocode}

\usepackage{amsmath, bm}

\usepackage[colorinlistoftodos,prependcaption,textsize=tiny]{todonotes}
\usepackage{lipsum}                     
\usepackage{xargs}                      
\newcommandx{\unsure}[2][1=]{\todo[linecolor=red,backgroundcolor=red!25,bordercolor=red,#1]{#2}}
\newcommandx{\change}[2][1=]{\todo[linecolor=blue,backgroundcolor=blue!25,bordercolor=blue,#1]{#2}}
\newcommandx{\info}[2][1=]{\todo[linecolor=green,backgroundcolor=green!25,bordercolor=green,#1]{#2}}
\newcommandx{\improvement}[2][1=]{\todo[linecolor=Plum,backgroundcolor=Plum!25,bordercolor=Plum,#1]{#2}}

\definecolor{TUMBlue}{HTML}{0065BD}
\definecolor{TUMSecondaryBlue}{HTML}{005293}
\definecolor{TUMSecondaryBlue2}{HTML}{003359}
\definecolor{TUMBlack}{HTML}{000000}
\definecolor{TUMWhite}{HTML}{FFFFFF}
\definecolor{TUMDarkGray}{HTML}{333333}
\definecolor{TUMGray}{HTML}{808080}
\definecolor{TUMLightGray}{HTML}{CCCCC6}
\definecolor{TUMAccentGray}{HTML}{DAD7CB}
\definecolor{TUMAccentOrange}{HTML}{E37222}
\definecolor{TUMAccentGreen}{HTML}{A2AD00}
\definecolor{TUMAccentLightBlue}{HTML}{98C6EA}
\definecolor{TUMAccentBlue}{HTML}{64A0C8}

\tikzstyle{arrow1} = [line width=0.2mm,->,>=stealth,rounded corners=4pt, draw=black!70, align=center]
\tikzstyle{arrow} = [thick,->,>=stealth,rounded corners=4pt, draw=black, align=center]
\tikzstyle{dashedarrow} = [thick,->,>=stealth,rounded corners=4pt, draw=black, align=center, dashed]
\tikzstyle{graydashedarrow} = [thick,->,>=stealth,rounded corners=4pt, draw=black, align=center, dashed, color=TUMGray]
\tikzstyle{orangedashedarrow} = [thick,->,>=stealth,rounded corners=4pt, draw=black, align=center, dashed, color=TUMAccentOrange]

\tikzstyle{marker} = [circle, minimum size=0.4cm, scale=0.8, text centered, draw=black, align=center, fill=orange!10, font=\large]
\tikzstyle{circlenum} = [circle, draw=blue,thick,inner sep=1.5mm]

\usepackage{pgfplots}
\usepackage{pgfplotstable}
\usepackage{tikz}
\tikzset{>=stealth}
\usetikzlibrary{patterns}
\usetikzlibrary{pgfplots.statistics}
\usetikzlibrary{positioning, shapes}
\usetikzlibrary{calc}
\usetikzlibrary{backgrounds,scopes}
\usetikzlibrary{fit}
\usetikzlibrary{arrows.meta}

\tikzstyle{neuron} = [circle, minimum size=0.5cm, text width=0.2cm, text height=0.2cm, text centered, draw=black, align=center, fill=TUMBlue!20, font=\footnotesize]
\tikzstyle{weight} = [thin,-,>=stealth, draw=black!80, align=center]

\tikzstyle{latent} = [rectangle, minimum width=1cm, minimum height=1cm, text centered, draw=black, rounded corners, align=center, fill=TUMAccentLightBlue]
\tikzstyle{loss} = [rectangle, minimum width=1cm, minimum height=1.2cm, text centered, draw=black, align=center]
\tikzstyle{module} = [rectangle, minimum width=2.5cm, minimum height=2cm, text centered, draw=black, align=center]
\tikzstyle{dark_module} = [rectangle, minimum width=2.5cm, minimum height=2cm, text centered, draw=black, rounded corners, align=center, fill=TUMAccentLightBlue]

\tikzstyle{arrow} = [thick,->,>=stealth,rounded corners=4pt, draw=black, align=center]
\tikzstyle{dashedarrow} = [thick,->,>=stealth,rounded corners=4pt, draw=black, align=center, dashed]
\tikzstyle{graydashedarrow} = [thick,->,>=stealth,rounded corners=4pt, draw=black, align=center, dashed, color=TUMGray]
\tikzstyle{orangedashedarrow} = [thick,->,>=stealth,rounded corners=4pt, draw=black, align=center, dashed, color=TUMAccentOrange]

\tikzstyle{marker} = [circle, minimum size=0.4cm, scale=0.8, text centered, draw=black, align=center, fill=orange!10, font=\large]
\tikzstyle{circlenum} = [circle, draw=blue,thick,inner sep=1.5mm]

\newcommand{\drawMLTen}
{
	\centering
	\begin{tikzpicture} [scale=0.8, transform shape]
		\centering
		\begin{axis}[
			ybar, axis on top,
			height=7cm, width=17cm,
			bar width=0.1cm,
			ymajorgrids, tick align=inside,
			major grid style={draw=gray!50,dashed},
			enlarge y limits={value=.1,upper},
			ymin=0, ymax=100,
			axis x line*=bottom,
			axis y line*=right,
			y axis line style={opacity=0},
			tickwidth=0pt,
			enlarge x limits=0.02,
			legend style={
				at={(0.5,1.1)},
				anchor=north,
				legend columns=-1,
				/tikz/every even column/.append style={column sep=0.5cm}
			},
			ylabel={Averaged Success Rate (\%)},
			symbolic x coords={
				\textbf{train average},reach,press button, open door,close drawer,push, basketball, open window, pick and place, 
				\textbf{test average}, open drawer, close door, sweep into goal, place onto shelf, pull lever},
			xtick=data, x tick label style={rotate=45,anchor=east},
			]

			\addplot [draw=none, fill=blue!50] coordinates {
				(\textbf{train average},98.8)
				(reach, 99.296) 
				(press button,100)
				(open door,100) 
				(close drawer,100)
				(push,100) 
				(basketball,100) 
				(open window,97.8) 
				(pick and place,100) 
				(\textbf{test average},50) 
				(open drawer,0) 
				(close door,83.33) 
				(sweep into goal,100) 
				(place onto shelf,62.222) 
				(pull lever,0) 
			};
			\addplot [draw=none,fill=red!50] coordinates {
				(\textbf{train average},46.102)
				(reach, 100) 
				(press button,100)
				(open door,29.830) 
				(close drawer,44.068)
				(push,0) 
				(basketball,0) 
				(open window,0) 
				(pick and place,10.847) 
				(\textbf{test average},0) 
				(open drawer,0) 
				(close door,0) 
				(sweep into goal,0) 
				(place onto shelf,0) 
				(pull lever,0) 
			};
			\addplot [draw=none, fill=green!50] coordinates {
				(\textbf{train average},50.169)
				(reach, 69.830) 
				(press button,100)
				(open door,60) 
				(close drawer,50.169)
				(push,29.830) 
				(basketball,29.830) 
				(open window,20) 
				(pick and place,20) 
				(\textbf{test average},10.169) 
				(open drawer,0) 
				(close door,0) 
				(sweep into goal,29.830) 
				(place onto shelf,20) 
				(pull lever,0) 
			};
			\addplot [draw=none, fill=yellow!50] coordinates {
				(\textbf{train average},25.084)
				(reach, 20) 
				(press button,90)
				(open door,50.169) 
				(close drawer,0)
				(push,40) 
				(basketball,0) 
				(open window,29.830) 
				(pick and place,20) 
				(\textbf{test average},35.254) 
				(open drawer,1) 
				(close door, 50) 
				(sweep into goal,29.830) 
				(place onto shelf,0) 
				(pull lever,0) 
			};
			\draw [line width=0.5mm, red, dashed] (85, 0.1) -- (85, 100.2);
			
			\legend{MILLION, PEARL, RL$^2$-PPO, MAML-TRPO}
		\end{axis}
	\end{tikzpicture}
}

\newcommand{\drawMLTenV}
{
	\centering
	\begin{tikzpicture} [scale=0.8, transform shape]
		\centering
		\begin{axis}[
			ybar, axis on top,
			height=7cm, width=19cm,
			bar width=0.1cm,
			ymajorgrids, tick align=inside,
			major grid style={draw=gray!50,dashed},
			enlarge y limits={value=.1,upper},
			ymin=0, ymax=100,
			axis x line*=bottom,
			axis y line*=right,
			y axis line style={opacity=0},
			tickwidth=0pt,
			enlarge x limits=0.02,
			legend style={
				at={(0.5,1.1)},
				anchor=north,
				legend columns=-1,
				/tikz/every even column/.append style={column sep=0.5cm}
			},
			ylabel={Averaged Success Rate (\%)},
			symbolic x coords={
				\textbf{train average},reach,press button,sweep, open door,close drawer,push, basketball, open window, pick and place, insert peg,  
				\textbf{test average}, open drawer, close door, sweep into goal, place onto shelf, pull lever},
			xtick=data, x tick label style={rotate=45,anchor=east},
			]

			\addplot [draw=none, fill=blue!50] coordinates {
				(\textbf{train average},98.3)
				(reach, 98.2) 
				(press button,100)
				(sweep,100) 
				(open door,100) 
				(close drawer,100)
				(push,100) 
				(basketball,100) 
				(open window,98) 
				(pick and place,100) 
				(insert peg,98.2) 
				(\textbf{test average},55.43) 
				(open drawer,8.5) 
				(close door,100) 
				(sweep into goal,100) 
				(place onto shelf,35.68) 
				(pull lever,12.49) 
			};
			\addplot [draw=none,fill=red!50] coordinates {
				(\textbf{train average},28)
				(reach, 39) 
				(press button,45)
				(sweep,0) 
				(open door,0) 
				(close drawer,100)
				(push,7) 
				(basketball,0) 
				(open window,31) 
				(pick and place,0) 
				(insert peg,0) 
				(\textbf{test average},13) 
				(open drawer,0) 
				(close door,63) 
				(sweep into goal,20) 
				(place onto shelf,0) 
				(pull lever,20) 
			};
			\addplot [draw=none, fill=green!50] coordinates {
				(\textbf{train average},88)
				(reach, 100) 
				(press button,100)
				(sweep,94) 
				(open door,100) 
				(close drawer,100)
				(push,100) 
				(basketball,87) 
				(open window,100) 
				(pick and place,100)
				(insert peg,97) 
				(\textbf{test average},38) 
				(open drawer,23) 
				(close door,96) 
				(sweep into goal,91) 
				(place onto shelf,5) 
				(pull lever,17) 
			};
			\addplot [draw=none, fill=yellow!50] coordinates {
				(\textbf{train average},48)
				(reach, 100) 
				(press button,100)
				(sweep,2) 
				(open door,78) 
				(close drawer,100)
				(push,1) 
				(basketball,0) 
				(open window,88) 
				(pick and place,0) 
				(insert peg,0) 
				(\textbf{test average},33) 
				(open drawer,41) 
				(close door, 98) 
				(sweep into goal,6) 
				(place onto shelf,0) 
				(pull lever,37.5) 
			};
			\draw[line width=0.5mm, red, dashed] (105, 0.1) -- (105, 100.2);
			
			\legend{MILLION, PEARL, RL$^2$-PPO, MAML-TRPO}
		\end{axis}
	\end{tikzpicture}
}

\newcommand{\drawoverview}
{
	\centering
	\begin{tikzpicture} [scale=0.7, transform shape]
		\node(language) [module, minimum width=4cm, minimum height=1cm, fill=blue!10] at (-3, -0.2) {Language instructions};
		
		\node(glove) [module, minimum width=4cm, minimum height=1cm, fill=red!10] at (-3,1.3) {GloVe};
		
		\draw [arrow] (language) -- (glove);
		
		\node(transformer_1) [module, minimum width=2cm, minimum height=0.75cm] at (-3,3.4) {GTrXL};
		
		\draw [line width=0.5mm, dotted] (-2.75,4.05) -- (-3.25,4.05);
		
		\node(transformer_2) [module, minimum width=2cm, minimum height=0.75cm] at (-3,4.8) {GTrXL};
		
		\node(mlp_policy) [module, minimum width=1.75cm, minimum height=0.75cm, fill=yellow!10] at (-4,6.5) {MLP};
		
		\node(mlp_value) [module, minimum width=1.75cm, minimum height=0.75cm, fill=yellow!10] at (-2,6.5) {MLP};
		
		\node at (-4, 7.5) {Policy $\Pi_{\theta}$};
		\node at (-2, 7.5) {Value $V_{\phi}$};
		
		\draw [arrow] (mlp_policy) -- (-4, 7.25);
		\draw [arrow] (mlp_value) -- (-2, 7.25);

		\begin{scope}[on background layer]
			\node [rectangle, draw, text centered, rounded corners, minimum width=3.5cm, minimum height=6.5cm, fit={(transformer_1) (transformer_2) (glove) (language) (mlp_policy) (mlp_value)}, fill=black!10] (transformer_full) at (-3, 3.45) {};
		\end{scope}
		
		\begin{scope}[on background layer]
			\node [rectangle, draw, text centered, rounded corners, minimum width=2.9cm, minimum height=2.25cm, fit={(transformer_1) (transformer_2)}, fill=green!20] (transformer) at (-3, 4.05) {};
			\node[anchor=south east] (phase1T) at (transformer.south east) {\textbf{Transformer}};
		\end{scope}
		
		\draw [arrow] (glove) -- (transformer);
		\draw [arrow] (transformer) -- (mlp_policy);
		\draw [arrow] (transformer) -- (mlp_value);
		
		\node(o_0) [neuron] at (1, 1.5) {$o_0$};
		\node(a_0) [neuron] at (2, 1.5) {$a_0$};
		\node(o_1) [neuron] at (3, 1.5) {$o_1$};
		\node(a_1) [neuron] at (4, 1.5) {$a_1$};
		\node(o_2) [neuron] at (5, 1.5) {$o_2$};
		\node(a_2) [neuron] at (6, 1.5) {$a_2$};
		\node(o_3) [neuron] at (7, 1.5) {$o_3$};
		\node(a_3) [neuron] at (8, 1.5) {$a_3$};
		\node(o_4) [neuron] at (9, 1.5) {$o_4$};
		\node(a_4) [neuron] at (10, 1.5) {$a_4$};
		\node(o_5) [neuron] at (11, 1.5) {$o_5$};
		\node(a_5) [neuron] at (12, 1.5) {$a_5$};
		\node(o_6) [neuron] at (13, 1.5) {$o_6$};
		\node(a_6) [neuron] at (14, 1.5) {$a_6$};
		\node(o_7) [neuron] at (15, 1.5) {$o_7$};
		\node(a_7) [neuron] at (16, 1.5) {$a_7$};
		
		\node(E_1_0) at (1, 3.5) {$E^1_0$};
		\node(E_2_0) at (1, 5.5) {$E^2_0$};
		\node(pi_0) at (1, 7.5) {$\pi(a_0|o_{0:0})$};
		\node(v_0) at (0.5, 6.75) {$V$};
		
		\draw [arrow] (o_0) -- (E_1_0);
		\draw [arrow] (E_1_0) -- (E_2_0);
		\draw [arrow] (E_2_0) -- (v_0);
		\draw [arrow] (E_2_0) -- (pi_0);
		\draw[->, dashed] (0, 3) -- (E_1_0);
		\draw[->, dashed] (0, 2.5) -- (E_1_0);
		\draw[->, dashed] (0, 5) -- (E_2_0);
		\draw[->, dashed] (0, 4.5) -- (E_2_0);
		
		\node(E_1_1) at (3, 3.5) {$E^1_1$};
		\node(E_2_1) at (3, 5.5) {$E^2_1$};
		\node(pi_1) at (3, 7.5) {$\pi(a_1|o_{0:1})$};
		\node(v_1) at (2.5, 6.75) {$V$};
		
		\draw [arrow] (o_1) -- (E_1_1);
		\draw [arrow] (o_0) -- (E_1_1);
		\draw [arrow] (E_1_0) -- (E_2_1);
		\draw [arrow] (E_1_1) -- (E_2_1);
		\draw [arrow] (E_2_1) -- (pi_1);
		\draw [arrow] (E_2_1) -- (v_1);
		\draw[->, dashed] (0, 2.0) -- (E_1_1);
		\draw[->, dashed] (0, 2.25) -- (E_1_1);
		\draw[->, dashed] (0, 4) -- (E_2_1);
		\draw[->, dashed] (0, 4.25) -- (E_2_1);
		
		\node(E_1_2) at (5, 3.5) {$E^1_2$};
		\node(E_2_2) at (5, 5.5) {$E^2_2$};
		\node(pi_2) at (5, 7.5) {$\pi(a_2|o_{0:2})$};
		\node(v_2) at (4.5, 6.75) {$V$};
		
		\draw [arrow] (o_2) -- (E_1_2);
		\draw [->, dashed] (o_0) -- (E_1_2);
		\draw [->, dashed] (o_1) -- (E_1_2);
		\draw [arrow] (E_1_2) -- (E_2_2);
		\draw [arrow] (E_2_2) -- (pi_2);
		\draw [arrow] (E_2_2) -- (v_2);
		\draw [->, dashed] (E_1_1) -- (E_2_2);
		\draw [->, dashed] (E_1_0) -- (E_2_2);
		
		\node(E_1_3) at (7, 3.5) {$E^1_3$};
		\node(E_2_3) at (7, 5.5) {$E^2_3$};
		\node(pi_3) at (7, 7.5) {$\pi(a_3|o_{0:3})$};
		\node(v_3) at (6.5, 6.75) {$V$};
		
		\draw [arrow] (o_3) -- (E_1_3);
		\draw [->, dashed] (o_0) -- (E_1_3);
		\draw [->, dashed] (o_0) -- (E_1_3);
		\draw [->, dashed] (o_1) -- (E_1_3);
		\draw [arrow] (o_2) -- (E_1_3);
		\draw [arrow] (E_1_3) -- (E_2_3);
		\draw [arrow] (E_2_3) -- (pi_3);
		\draw [arrow] (E_2_3) -- (v_3);
		\draw [arrow] (E_1_2) -- (E_2_3);
		\draw [->, dashed] (E_1_1) -- (E_2_3);
		\draw [->, dashed] (E_1_0) -- (E_2_3);
		
		\node(E_1_4) at (9, 3.5) {$E^1_4$};
		\node(E_2_4) at (9, 5.5) {$E^2_4$};
		\node(pi_4) at (9, 7.5) {$\pi(a_4|o_{0:4})$};
		\node(v_4) at (8.5, 6.75) {$V$};
		
		\draw [arrow] (o_4) -- (E_1_4);
		\draw [arrow] (E_1_4) -- (E_2_4);
		\draw [arrow] (E_2_4) -- (pi_4);
		\draw [arrow] (E_2_4) -- (v_4);
		\draw [->, dashed] (E_1_3) -- (E_2_4);
		\draw [->, dashed] (o_2) -- (E_1_4);
		\draw [->, dashed] (o_3) -- (E_1_4);
		\draw [->, dashed] (E_1_2) -- (E_2_4);
		
		\node(E_1_5) at (11, 3.5) {$E^1_5$};
		\node(E_2_5) at (11, 5.5) {$E^2_5$};
		\node(pi_5) at (11, 7.5) {$\pi(a_5|o_{0:5})$};
		\node(v_5) at (10.5, 6.75) {$V$};
		
		\draw [arrow] (o_5) -- (E_1_5);
		\draw [arrow] (E_1_5) -- (E_2_5);
		\draw [arrow] (E_2_5) -- (pi_5);
		\draw [arrow] (E_2_5) -- (v_5);
		\draw [arrow] (E_1_4) -- (E_2_5);
		\draw [->, dashed] (o_2) -- (E_1_5);
		\draw [->, dashed] (o_3) -- (E_1_5);
		\draw [arrow] (o_4) -- (E_1_5);
		\draw [->, dashed] (E_1_2) -- (E_2_5);
		\draw [->, dashed] (E_1_3) -- (E_2_5);
		
		\node(E_1_6) at (13, 3.5) {$E^1_6$};
		\node(E_2_6) at (13, 5.5) {$E^2_6$};
		\node(pi_6) at (13, 7.5) {$\pi(a_6|o_{0:6})$};
		\node(v_6) at (12.5, 6.75) {$V$};
		
		\draw [arrow] (o_6) -- (E_1_6);
		\draw [arrow] (E_1_6) -- (E_2_6);
		\draw [arrow] (E_2_6) -- (pi_6);
		\draw [arrow] (E_2_6) -- (v_6);
		\draw [->, dashed] (E_1_5) -- (E_2_6);
		\draw [->, dashed] (o_4) -- (E_1_6);
		\draw [->, dashed] (o_5) -- (E_1_6);
		\draw [->, dashed] (E_1_4) -- (E_2_6);
		\draw [->, dashed] (E_1_5) -- (E_2_6);
		
		\node(E_1_7) at (15, 3.5) {$E^1_7$};
		\node(E_2_7) at (15, 5.5) {$E^2_7$};
		\node(pi_7) at (15, 7.5) {$\pi(a_7|o_{0:7})$};
		\node(v_7) at (14.5, 6.75) {$V$};
		
		\draw [arrow] (o_7) -- (E_1_7);
		\draw [arrow] (E_1_7) -- (E_2_7);
		\draw [arrow] (E_2_7) -- (pi_7);
		\draw [arrow] (E_2_7) -- (v_7);
		\draw [arrow] (E_1_6) -- (E_2_7);
		\draw [->, dashed] (o_4) -- (E_1_7);
		\draw [->, dashed] (o_5) -- (E_1_7);
		\draw [arrow] (o_6) -- (E_1_7);
		\draw [->, dashed] (E_1_4) -- (E_2_7);
		\draw [->, dashed] (E_1_5) -- (E_2_7);
		
		\node(instruction_1) [module, minimum width=4cm, minimum height=3cm] at (2.5, 0.5) {};
		\node(trial_1) [module, minimum width=6cm, minimum height=3cm] at (7.5, 0.5) {};
		\node(instruction_2) [module, minimum width=4cm, minimum height=3cm] at (12.5, 0.5) {};
		
		\node(s_0) [neuron] at (5, 0.5) {$s_0$};
		\node(s_1) [neuron] at (7, 0.5) {$s_1$};
		\node(s_2) [neuron] at (9, 0.5) {$s_2$};
		\node(r_2) [neuron] at (6, -0.5) {$r_2$};
		\node(r_3) [neuron] at (8, -0.5) {$r_3$};
		\node(s_0_1) [neuron] at (15, 0.5) {$s_0$};
		\node(r_7) [neuron] at (16, -0.25) {$r_7$};
		
		\draw[arrow] (s_0) -- (o_2);
		\draw[arrow] (s_0) -- (s_1);
		\draw[arrow] (s_0) -- (r_2);
		\draw[arrow] (a_2) -- (r_2);
		\draw[arrow] (a_2) -- (s_1);
		\draw[arrow] (s_1) -- (o_3);
		\draw[arrow] (s_1) -- (r_3);
		\draw[arrow] (s_1) -- (s_2);
		\draw[arrow] (a_3) -- (r_3);
		\draw[arrow] (a_3) -- (s_2);
		\draw[arrow] (s_2) -- (o_4);
		\draw[arrow] (s_0_1) -- (o_7);
		\draw[arrow] (s_0_1) -- (r_7);
		\draw[arrow] (a_7) -- (r_7);
		\draw[arrow] (s_0_1) -- (16.5, 0.5);
		\draw[arrow] (a_7) -- (16.5, 1);
		
		\node at (3, -0.75) {\textbf{Instruction phase}};
		\node at (9.5, -0.75) {\textbf{Trial phase}};
		\node at (13, -0.75) {\textbf{Instruction phase}};
		\node at (15.5, -0.75) {\textbf{Trial phase}};
		\node at (1.5, 0.5) {\textbf{$r_0=0$}};
		\node at (3.5, 0.5) {\textbf{$r_1=0$}};
		\node at (11.5, 0.5) {\textbf{$r_5=0$}};
		\node at (13.5, 0.5) {\textbf{$r_6=0$}};
		\draw (14.5, 2) -- (16.5, 2);
		\draw (14.5, -1) -- (16.5, -1);
		\node at (9.5, -0.25) {$r_4=0$};
		
		\draw (9,0.5) circle (0.275cm);
		
		\begin{scope}[on background layer]
			\node [rectangle, dashed, draw, text centered, rounded corners, minimum height = 5cm, minimum width=0.05cm, fit={(o_0) (E_1_0) (E_2_0) (pi_0)}, fill=black!10] (transformer_simple)  {};
		\end{scope}
		\draw [->, black!50, ultra thick] (transformer_simple.north) to [out=135,in=45] (transformer_full.north);
		
	\end{tikzpicture}
}

\newcommand{\drawlanginstruction}
{
	\centering
	\begin{tikzpicture} [scale=0.7, transform shape]
		\draw [line width=0.5mm] (0, 0.25) -- (11, 0.25);
		\draw [line width=0.5mm] (0, 5) -- (11, 5);
		\draw [line width=0.5mm] (4.75, 0.25) -- (4.75, 5);
		
		\node[color=red!60] at (2.4, 0.75) {\textbf{Instruction phase}};
		
		\node(word_1) at (2.4, 1.5) {``the"};
		\node(w_1) at (2.4, 3) {$w_1 \in \mathbb{R}^{50}$};
		\node(o_1) at (2.4, 4.5) {$o_1 \in \mathbb{R}^{50}$};
		
		\draw[arrow] (word_1) -- (w_1);
		\draw[arrow] (w_1) -- (o_1);
		
		\node(word_0) at (0.9, 1.5) {``close"};
		\node(w_0) at (0.9, 3) {$w_0 \in \mathbb{R}^{50}$};
		\node(o_0) at (0.9, 4.5) {$o_0 \in \mathbb{R}^{50}$};
		
		\draw[arrow] (word_0) -- (w_0);
		\draw[arrow] (w_0) -- (o_0);
		
		\node(word_2) at (4, 1.5) {``drawer"};
		\node(w_2) at (4, 3) {$w_2 \in \mathbb{R}^{50}$};
		\node(o_2) at (4, 4.5) {$o_2 \in \mathbb{R}^{50}$};
		
		\draw[arrow] (word_2) -- (w_2);
		\draw[arrow] (w_2) -- (o_2);
		
		\node[color=blue!60] at (8.1, 0.75) {\textbf{Trial phase}};
		
		\node(word_3) at (6.0, 1.5) {Joint positions};
		\node(w_3) at (6.0, 3) {$s_0 \in \mathbb{R}^{6}$};
		\node(o_3) at (6.0, 4.5) {$o_3 \in \mathbb{R}^{50}$};
		
		\draw[arrow] (word_3) -- (w_3);
		\draw[arrow] (w_3) -- (o_3);
		
		\node at (7.4, 4) {\textbf{concat($s_0$, [0,...])}};
		
		\node(word_4) at (9.0, 1.5) {Joint positions};
		\node(w_4) at (9.0, 3) {$s_1 \in \mathbb{R}^{6}$};
		\node(o_4) at (9.0, 4.5) {$o_4 \in \mathbb{R}^{50}$};
		
		\draw[arrow] (word_4) -- (w_4);
		\draw[arrow] (w_4) -- (o_4);
		
		\node at (10.4, 4) {\textbf{concat($s_0$, [0,...])}};
		
		\node(reset) at (4.75, 6) {\textbf{Env. Reset}};
		\draw[arrow] (reset) -- (4.75, 5);
	\end{tikzpicture}
}

\newcommand{\drawtrialphase}
{
	\centering
	\begin{tikzpicture} [scale=0.7, transform shape]
		\draw [line width=0.5mm] (1, 0) -- (11, 0);
		\draw [line width=0.5mm] (1, 1.5) -- (11, 1.5);
		\draw [line width=0.5mm] (1, 0) -- (1, 1.5);
		\draw [line width=0.5mm] (3.5, 0) -- (3.5, 1.5);
		\draw [line width=0.5mm] (6, 0) -- (6, 1.5);
		\draw [line width=0.5mm] (8.5, 0) -- (8.5, 1.5);
		\draw [line width=0.5mm] (11, 0) -- (11, 1.5);
		
		\node(word_0) at (2.25, 0.75) {Instruction};
		\node(word_0) at (4.75, 0.75) {Trial};
		\node(word_0) at (7.25, 0.75) {Trial};
		\node(word_0) at (9.75, 0.75) {Trial};
		
		\node(word_0) at (0, 0.75) {\boldsymbol{$\pi(a_t|o_{0:t})$}};
		\node at (1, 2.25) {\textbf{Episode starts}};
		\draw[arrow] (1, 2) -- (1, 1.5);
		
		\draw [decorate,decoration={brace,amplitude=15pt}]
		(3.5,1.6) -- (11,1.6) node [midway, yshift=0.4in] {\textbf{Three trials}};
		
		\node at (2.25, -0.75) {\textbf{``Fold the towel"}};
		
		\node at (11, -0.75) {\textbf{Episode ends}};
		\draw[arrow] (11, -0.5) -- (11, 0);
		
		\node at (1, 3) {\textbf{\textcolor{red}{An successful trial}}};

		\draw [line width=0.5mm] (1, -5) -- (11, -5);
		\draw [line width=0.5mm] (1, -3.5) -- (11, -3.5);
		\draw [line width=0.5mm] (1, -5) -- (1, -3.5);
		\draw [line width=0.5mm] (3.5, -5) -- (3.5, -3.5);
		\draw [line width=0.5mm] (6, -5) -- (6, -3.5);
		\draw [line width=0.5mm] (8.5, -5) -- (8.5, -3.5);
		\draw [line width=0.5mm] (11, -5) -- (11, -3.5);
		
		\node(word_0) at (2.25, -4.25) {Instruction};
		\node(word_0) at (4.75, -4.25) {Trial};
		\node(word_0) at (7.25, -4.25) {Instruction};
		\node(word_0) at (9.75, -4.25) {Trial};
		
		\node(word_0) at (0, -4.25) {\boldsymbol{$\pi(a_t|o_{0:t})$}};
		\node at (1, -2.75) {\textbf{Episode starts}};
		\draw[arrow] (1, -3) -- (1, -3.5);
		
		\node at (2.25, -5.75) {\textbf{``Fold the towel"}};
		
		\node at (7.25, -5.75) {\textbf{``Fold the towel twice"}};
		
		\node at (11.5, -4.25){\textbf{...}};
		
		\node at (1, -2) {\textbf{\textcolor{red}{An unsuccessful trial}}};

	\end{tikzpicture}
}

\renewcommand{\baselinestretch}{0.99}  
\begin{document}
%
\title{Meta-Reinforcement Learning via Language Instructions}
%
%
%

\author{Zhenshan~Bing,
        Alexander~Koch,
        Xiangtong~Yao,
        Kai~Huang,
        Alois~Knoll
\thanks{Z. Bing, A. Koch, X. Yao, and A. Knoll are with the Department
of Informatics, Technical University of Munich, Germany.
E-mail: \{bing, yaox, knoll\}@in.tum.de}
\thanks{K. Huang is with the School of Data and Computer Science, Sun Yat-sen University, China. Email: huangk36@mail.sysu.edu.cn}
}

%
%

\markboth{Journal of \LaTeX\ Class Files,~Vol.~14, No.~8, August~2015}%
{Shell \MakeLowercase{\textit{et al.}}: Bare Demo of IEEEtran.cls for IEEE Journals}
%



\maketitle

\begin{abstract}
Although deep reinforcement learning has recently been very successful at learning complex behaviors, it requires a tremendous amount of data to learn a task.
One of the fundamental reasons causing this limitation lies in the nature of the trial-and-error learning paradigm of reinforcement learning, where the agent communicates with the environment and progresses in the learning only relying on the reward signal.
This is implicit and rather insufficient to learn a task well. 
On the contrary, humans are usually taught new skills via natural language instructions.
Utilizing language instructions for robotic motion control to improve the adaptability is a recently emerged topic and challenging.
In this paper, we present a meta-RL algorithm that addresses the challenge of learning skills with  language instructions in multiple manipulation tasks.
On the one hand, our algorithm utilizes the language instructions to shape its interpretation of the task, on the other hand, it still learns to solve task in a trial-and-error process.
We evaluate our algorithm on the robotic manipulation benchmark (Meta-World) and it significantly outperforms state-of-the-art methods in terms of training and testing task success rates.
Codes are available at \url{https://tumi6robot.wixsite.com/million}.
\end{abstract}


%
\IEEEpeerreviewmaketitle

\section{Introduction}%

In recent years, deep reinforcement learning (RL) has been applied very successfully to hard control tasks like playing video games \cite{Vinyals2019, Berner2019, Mnih2015, Silver2017}, acquiring locomotion skills \cite{Schulman2015a, Haarnoja2018a, Barth-Maron2018} and, robotic manipulation tasks \cite{Levine2016, Andrychowicz2017, Akkaya2019}. 
However, learning these tasks often requires enormous amounts of environment interactions, which makes it impractical for many applications. 
For example, learning to manipulate a Rubik's cube for a robotic hand, OpenAI reported a cumulative experience of 14,000 years for simulated interactions \cite{Akkaya2019}. 
On the contrary, humans are able to manipulate the cube nearly instantaneously, as they have learned how to manipulate objects in general beforehand.

Meta-reinforcement learning (meta-RL) aims to design an efficient reinforcement learning algorithm to mimic the human learning ability that learns new tasks quickly \cite{Duan2016, Wang2016, Botvinick2019}. 
Meta-RL algorithms achieve this by conditioning the policy on past experience and inferring the task information based on the received rewards \cite{Rakelly2019}. 
Unfortunately, meta-RL algorithms perform poorly on diverse sets of tasks \cite{Yu2019a}, since they solely rely on rewards to communicate the task to the agent, which is especially problematic when the rewards are sparse or indistinguishable among similar tasks. 
Therefore, providing additional information about the task to the agent offers a promising way to help the learning of new tasks.
Natural language provides a rich and intuitive way for humans and robots to interact with each other, due to the possibility of referring to abstract concepts. 
When a human worker is given a new task, they are usually told what to do by language, which specifies the task goal or the required skill.
Therefore, the worker will not have to try every possible action sequence to figure out the goal, but purposefully aim at solving the specified task.
Although language is the most intuitive way for humans to understand tasks, the topic of controlling a robot using language instructions is rather new and poorly understood. 

\begin{figure}[t!]
	\centering
	\begin{tikzpicture}[scale=1]
		
		\node[inner sep=0pt,opacity=1] (base) at (0,0)
		{\includegraphics[width=250pt]{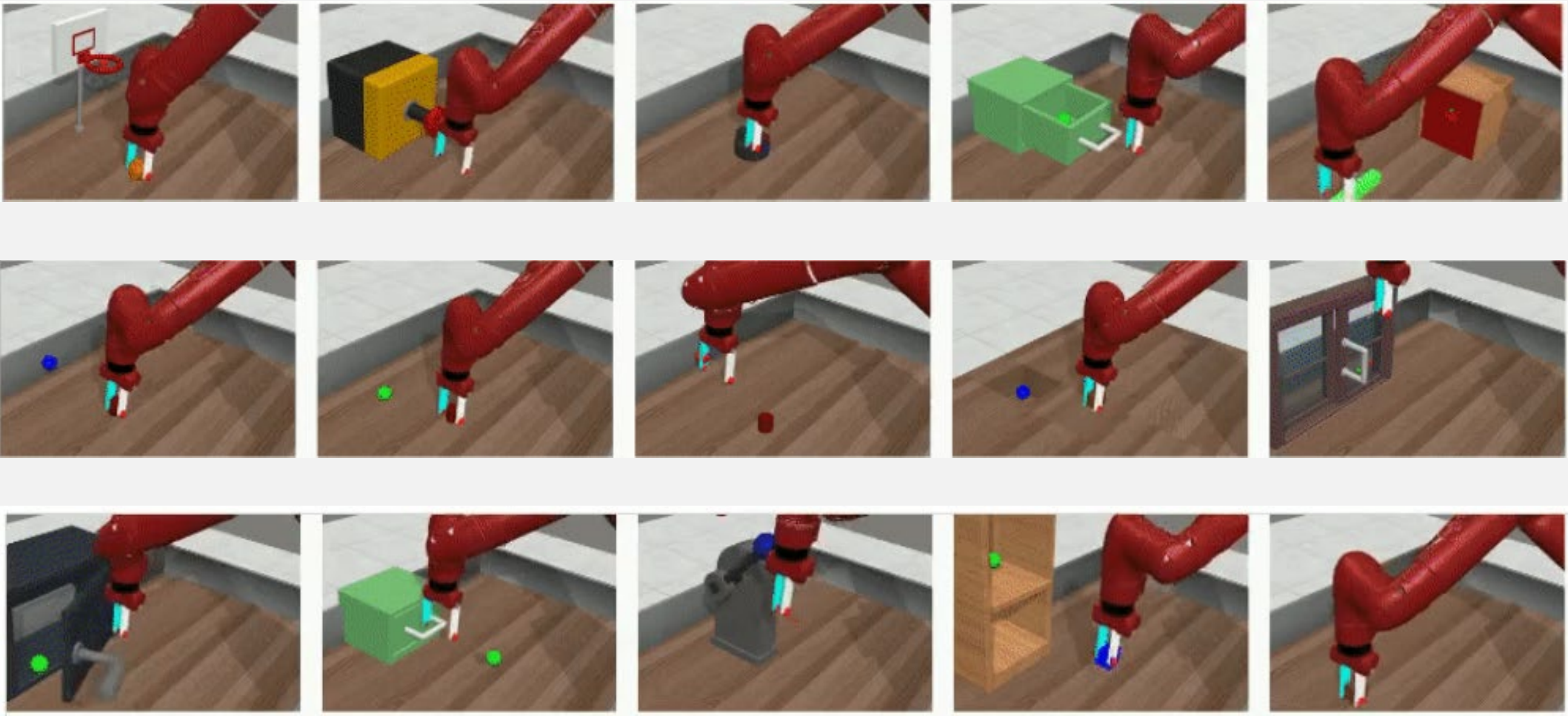}};
		
		\begin{scope}[xshift=0cm, yshift=0cm]
			\node[text=red!60, font=\footnotesize] at (-3.5, 0.7) {Basketball};
			\node[text=red!60, font=\footnotesize] at (-1.8, 0.7) {Button Press};
			\node[text=red!60, font=\footnotesize] at (-0.1, 0.7) {Dial Turn};
			\node[text=red!60, font=\footnotesize] at (1.7, 0.7) {Drawer Close};
			\node[text=red!60, font=\footnotesize] at (3.5, 0.7) {Peg Insert};
			
			\node[text=red!60, font=\footnotesize] at (-3.5, -0.7) {Pick Place};
			\node[text=red!60, font=\footnotesize] at (-1.8, -0.7) {Push};
			\node[text=red!60, font=\footnotesize] at (-0.1, -0.7) {Reach};
			\node[text=red!60, font=\footnotesize] at (1.7, -0.7) {Sweep Into};
			\node[text=red!60, font=\footnotesize] at (3.5, -0.7) {Window Open};
			
			\node[text=blue!60, font=\footnotesize] at (-3.5, -2.2) {Door Close};
			\node[text=blue!60, font=\footnotesize] at (-1.8, -2.2) {Drawer Open};
			\node[text=blue!60, font=\footnotesize] at (-0.1, -2.2) {Lever Pull};
			\node[text=blue!60, font=\footnotesize] at (1.7, -2.2) {Shelf Place};
			\node[text=blue!60, font=\footnotesize] at (3.5, -2.2) {Sweep};
			
		\end{scope}
		
	\end{tikzpicture}
	\vspace{-0.5cm}
	\caption{A visualization of the ML10 benchmark from Meta-World.
		The first two rows show the training tasks and the last row shows the testing tasks.
		The figure is adapted from \cite{Yu2019a}.}
	\label{fig_metaworld}
\end{figure}
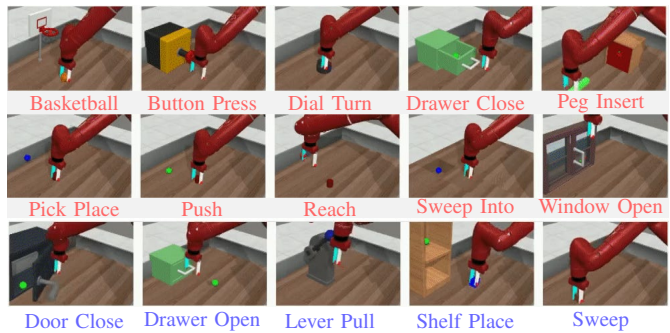

\begin{figure*}[!t]
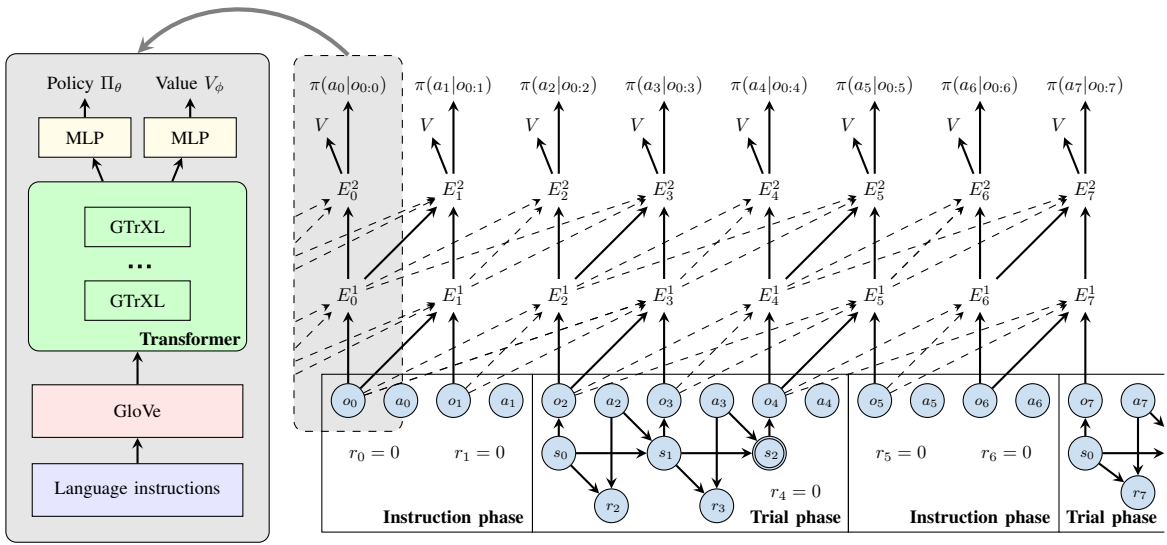

	\drawoverview
	\caption{Overview of our algorithm. Actions $a_t$ are sampled from the distribution $\pi(a_t|o_{0:t})$. The dotted lines indicate how the memory segment in the GTrXL influences the hidden states. The memory segment before the first observation of the episode is initialized as a sequence of zero vectors. States with a double circle are terminal states. In our experiments we use a larger size of the GTrXL such that the agent can still use the observations from the first instruction phase to compute the last few actions of an episode.
	}
	\label{fig:overview}
\end{figure*}

With the fast development of algorithms in natural language processing, more and more studies that  attempt to control robots via language instructions are beginning to emerge.
Shao et al. proposed an imitation learning algorithm Concept2Robot \cite{Shao2020}, which aims to enable the robot to learn manipulation skills from language instructions and visual appearances of the task in two stages.
In the first stage, Concept2Robot uses a video-based action classifier to generate a prediction score of the corresponding target task, which is served as a proxy reward to train the single-task policy.
In the second stage, a multi-task policy is trained through imitation learning to imitate all the single-task policies.
Stepputtis et al. \cite{NEURIPS2020_9909794d} introduced an imitation learning model that directly maps labeled language instructions and visual observations to manipulation skills.
Brucker et al. \cite{bucker2022reshaping} proposed a flexible language based interface for human-robot collaboration, which allows a user to reshape existing trajectories for an autonomous agent.
On the basis of imitating a large number of existing trajectories, the agent can generalize and adapt to new trajectories guided by the language. 
Lynch et al. \cite{pmlr-v100-lynch20a} invented another algorithm that learns from existing expert demonstrations and adapt to solve tasks via multi-modal information to create the goal, such as languages or images. 
Clearly, we can see most existing algorithms learn language-conditioned skills via the concept of imitation learning, where large numbers of expert trajectories are required.
This once again highly involves hand-crafted or engineered data and lacks the advantage of the trial-and-error learning paradigm, with which the agent can explore and learn the task by itself.

To this end, we establish a meta-RL algorithm that addresses the challenge of learning skills with language instructions in multiple manipulation tasks.
We introduce the \textbf{M}eta re\textbf{I}nforcement \textbf{L}earning algorithm using \textbf{L}anguage \textbf{I}nstructi\textbf{ON} (\textbf{MILLION}), which mimics the human-like learning manner and greatly improves the asymptotic performance in the challenging benchmark Meta-World.
We base our method on three concepts.
\begin{itemize}
	\item First, we propose a meta-RL learning paradigm that contains an instruction phase and a trial phase.
	In the instruction phase, the language description of the task is given to the agent, so that it can understand the goal of the task.
	In the trial phase, with the stored task information, the agent can explore and attempt to solve the task as standard reinforcement learning.
	\item Second, we build the architecture of our algorithm via three functional modules. 
	The language instruction is encoded by a pre-trained language module and then taken as an input for a transformer module, where the information is stored and processed.
	The on-policy RL algorithm V-MPO is used to update the policy network and the value network. 
	\item Experiment results demonstrate that MILLION significantly outperforms state-of-the-art algorithms on the challenging robotic manipulation benchmark (Meta-World \cite{Yu2019a}, Figure \ref{fig_metaworld}), in terms of training and testing success rate.
	Previous works only achieve less than $50\%$ success rate on the training tasks and less than $40\%$ on the testing tasks, while MILLION achieves almost perfect performance on the training tasks and can solve about half of the testing tasks.
\end{itemize}

\section{Methodology}

In this work, our goal is to propose a method that can provide the task information to the agent via instructions and learns to solve the task using trial-and-error RL algorithms.  
First, our policy network should be able to accept free-form language instructions of tasks as the input.
Second, our method should use such instructions to communicate to the agent about what the task entails, instead of using extensive numbers of expert trajectories as other imitating learning based methods.
Third, our method should enable the agent to successfully master diverse skills across broad tasks during training and adapt to unseen tasks during testing.


\subsection{Overview}

The architecture of MILLION is shown in Figure \ref{fig:overview} and briefly explained as follows.
\begin{itemize}
	\item First, an episode starts with an instruction phase, during which the language instructions are encoded as the observation using the pre-trained language model GloVe \cite{pennington2014glove} and fed into the transformer module.
	The action generated by the policy network and the reward collected from the environment are simply ignored, since there is no interaction during the instruction phase.
	\item Second, after the instruction phase, a trial phase is started, during which the agent interacts with the environment by following the task's Markov decision process (MDP).
	If the agent solves the task successfully, the environment will be reset and another trial phase starts.
	In the case of an unsuccessful trial, another instruction phase will start right after the trial phase, which resembles a real world scenario where a human operator might try a slightly different instruction to communicate the task to the agent.
	The whole episode will be terminated after a fixed number of trial phases.
	\item Finally, a new task is sampled and the same procedure will be executed.
	
\end{itemize}

\subsection{Language Instruction Phase}

\textcolor{black}{We consider the problem of learning an instruction-conditioned control policy $\pi(a|s, \mathcal{I}(\tau))$, where $\mathcal{I}$ represents language instructions about task $\tau$.
	$a$ is the selected action conditioned on the observation $s$. 
	The instructions should be encoded into a sequence of vectors $[w_1, w_2, \ldots , w_n] = \mathcal{I}(\tau) \text{, where } w_i \in \mathbb{R}^n, n \in \mathbb{N} $.
	We assume two phases in one training episode, namely, the \textbf{instruction phase} during which the task information is provided to the agent and the \textbf{trial phase} during which the agent interacts with the environment.
	The additional task information $\mathcal{I}(\tau)$ is only given to the agent in instruction phases, which can be expressed as
	\begin{equation}
		\begin{cases}
			\mathcal{I}(\tau) = [w_1, w_2, \ldots , w_n]_{1 \times n}& \text{, if } \textbf{instruction phase} \\ 
			\mathcal{I}(\tau) = [0, \ldots , 0]_{1 \times n} & \text{, if } \textbf{trial phase} 
		\end{cases}
	\end{equation}
	During the instruction phase, the agent receives the encoded vectors in sequence and does not interact with the environment, thus the actions generated by the policy and the environment rewards are ignored. While in the trial phase, the agent does not receive new instructions, but interacts with the environment via the actions and rewards, therefore, $\mathcal{I}(\tau)$ is set as zero.
}
\begin{figure}[!t]
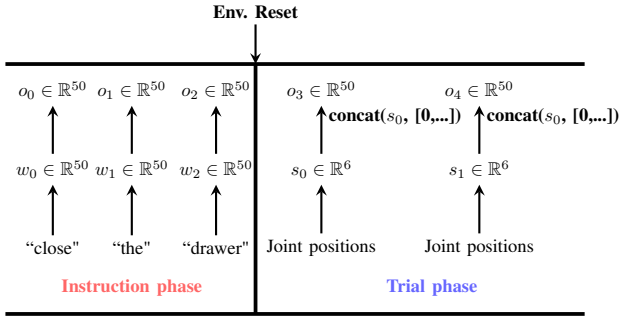

	\centering
	\drawlanginstruction
	\caption{Example of a language instruction during the instruction phase.
	}
	\label{fig:language_example}
\end{figure}

We provide free-form language instructions as the source of instructions, e.g., “open the drawer” for the drawer opening task, and “press the button” for the button pressing task.
For every task, we create a set of language instructions $\mathit{l}$ with similar key words.
Some examples of the language instructions that we use for the ML10 benchmark are listed in Table \ref{tab:language_configuration}.
At the beginning of the instruction phase, a new language instruction will be sampled for the current task $\tau$. 
To capture the information represented in the natural-language command, we first use the GloVe algorithm \cite{Pennington2014} to convert the language instruction $\mathit{l}$ into a sequence of fixed size vector $W = [w_0, ..., w_T] = \mathcal{I}(l)\text{ with } w_i \in \mathbb{R}^{50}$, encoding up to $T$ words with their respective 50-dimensional word embedding.
This means that, at time step $t$ of the instruction phase, the observation will be $w_t$. After $T$ time steps the trial phase will start. 
An example of the instruction phase is visualized in Figure \ref{fig:language_example}.
It should be noted that, to make the observations have the same length between the instruction phase and trial phase, a vector of zero is concatenated to the joint positions.

\begin{figure}[!t]
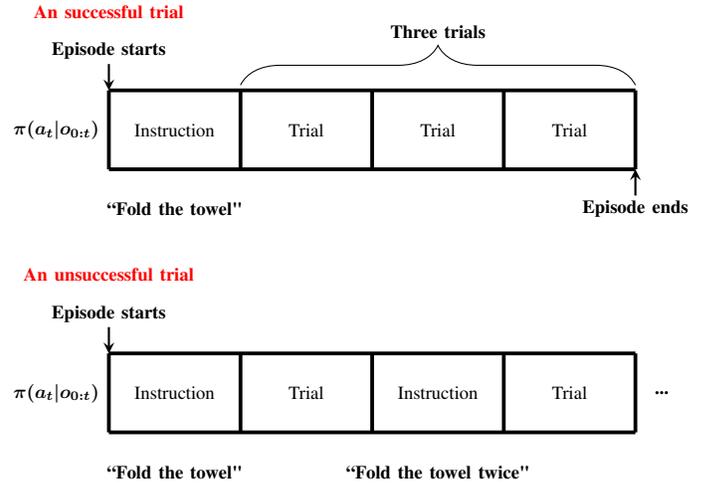

	\drawtrialphase
	\caption{The visualization of the phase sequence. Each episode starts with the instruction phase and follows with the trial phase.
		In case a trial is not able to solve the task, a new instruction phase will be added to enhance the understanding of the task.
	}
	\label{fig:trial_phase}
\end{figure}

\begin{table}[!b]   
	\centering
	\caption{Examples of Language instructions for ML10}
	\label{tab:language_configuration}
	\resizebox{0.95\columnwidth}{!}{
		\begin{tabular}{ll}
			\toprule
			\textbf{Task} & \textbf{Language Instructions} \\
			\hline
			reach & reach to goal\_pos, reach goal\_pos \\
			\hline
			\multirow{2}{*}{\makecell{push}} & {push goal\_pos, push to goal\_pos}  \\
			& {push object to goal\_pos} \\
			\hline
			\multirow{2}{*}{\makecell{pick-place}} & {pick and place at goal\_pos}  \\
			& {pick object and place at goal\_pos} \\
			\hline
			door-open  &  pull goal\_pos, open door, pull to goal\_pos \\
			\hline
			\multirow{2}{*}{\makecell{drawer-open}} & {pull goal\_pos, pull to goal\_pos}  \\
			& {pull back to goal\_pos} \\
			\hline
			\multirow{2}{*}{\makecell{drawer-close}} & {push goal\_pos, push to goal\_pos}  \\
			& {push forward to goal\_pos} \\
			\hline
			\multirow{2}{*}{\makecell{button-press-topdown}} & {push object down to goal\_pos, press button}  \\
			& {press down, press button down} \\
			\bottomrule
	\end{tabular}}
	\centering
\end{table}

\subsection{Trial Phase}

The trial phase is defined as steps of environmental interactions between two resets of the environment by following the task's MDP.
The action policy $\Pi$ and value policy $V$ are updated by maximizing the accumulated rewards in the trial phase.
The reset of the environment can be triggered by two conditions, namely, reaching a terminal state or reaching the maximum time-steps.
As illustrated in Figure \ref{fig:trial_phase}, we start each episode with an instruction phase and end the episode after three trial phases.
In the event of an successful trial in which the agent solves the task, we continue the training with a new trial phase.
In the event of an unsuccessful trial phase, we continue the training with a new instruction phase in which a similar language instruction is given to the agent.
Following the same procedure, one trial phase will be initiated after each instruction phase.

\subsection{Reward Normalization}
In multi-task RL or meta RL, one policy is trained to solve multiple tasks, from which the rewards typically have different magnitudes, for instance, in Meta-World (version 1), the task \textit{press-button-v1} has a reward varying from $0$ to $10,000$ while \textit{put-on-shelf-v1} has a reward varying from $0$ to $10$. 
This makes the learning extremely difficult and inefficient. 
A well-used solution is to clip the reward to a specified range. 

Preserving outputs precisely, while adaptively rescaling targets (Pop-Art) \cite{VanHasselt2016} can be used to
normalize the learning targets for the value function for every task individually.
Inspired by Pop-Art, we also update the value function of our network as follows.
The value function is used to predict the reward return $G_t$ and is approximated as
\begin{equation}
	f_{\theta, \sigma, \mu, w, b}(x) = \sigma (W h_{\theta / {W, b}} + b) + \mu,
\end{equation}
where $h$ is the neural network with the weights $\theta$. 
$W$ and $b$ are parameters to normalize the prediction of the network. 
$\mu$ and $\sigma$ are used to track the mean and standard deviation of the returns $G_t$.
Then, $\mu$ and $\sigma$ are updated as
\begin{equation}
	\begin{cases}
		\mu_t = (1 - \beta) \mu_{t-1} + \beta G_t &\\ 
		\sigma_t = \sqrt{\nu_t - \mu_t^2} &  \\ 
		\nu_t = (1 - \beta) \nu_{t-1} + \beta (G_t)^2 & 
	\end{cases}
\end{equation}
where $\beta$ is a training hyper-parameter.
To keep the learning stationary, we update $W$ and $b$ as 
\begin{equation}
	\begin{cases}
		W_t = \frac{\sigma_{t-1}}{\sigma_t} W_{t-1} &\\ 
		b_t = \frac{\sigma_t b_{t-1} + \mu_{t-1}  - \mu_t}{\sigma_t} & 
	\end{cases}. 
\end{equation}

\begin{algorithm}[!t]
	\caption{MILLION}
	\label{algo_million}
	\begin{algorithmic}[1]
		
		\State policy $\pi_\theta(a | s)$, state-value function $V_\phi^\pi(s) $
		\State initialize FIFO buffer $\tilde{B}$ with capacity $b * T_{target}$
		\While{not converged}
		\State Update $\pi_{\theta_{old}} \leftarrow \pi_\theta$
		\For {learning step $l=1..T_{target}$}
		\For {trajectory number $i=1...b$}
		\State Select instruction $I(\tau)$ for random task $\tau$
		\State Encode $I(\tau)$ in language phase
		\State Do MDPs in trial phase with $\pi_{\theta_{old}}(a|s,I(\tau))$ to generate trajectory $\Omega_{\tau}$, and add $\Omega_{\tau}$ to $\tilde{B}$
		\EndFor
		\State $B_{batch}$ = Sample $b$ trajectories from $\tilde{B}$
		\State Reward normalize $B_{batch}$
		\State Compute loss $\mathcal{L}(\phi, \theta, \eta, \alpha_\mu,\alpha_\Sigma)$ from $B_{batch}$
		\State Update $\phi, \theta, \eta, \alpha_\mu, \alpha_\Sigma$ with gradient step
		\EndFor
		\EndWhile
		
	\end{algorithmic}
\end{algorithm}

\subsection{V-MPO with Improved Sample Efficiency}

The policy is trained using the on-policy algorithm Maximum a Posteriori Policy Optimization (V-MPO).
V-MPO is very sample inefficient. 
It requires a lot of environment interactions during training. 
We improve the sample efficiency by modifying the V-MPO algorithm slightly to reuse sampled environment interactions more often. 
The original V-MPO algorithm uses every environment trajectory only for one gradient update. We change this by keeping a small FIFO buffer with the last $T_{target} \times b$ trajectories, where $b$ is the batch size for the gradient updates. 
Then we randomly sample batches from this buffer for gradient updates. 

The overall MILLION algorithm is given in Algorithm \ref{algo_million}.

\section{Experiments}
\label{chapter:experiments}

In this section, we evaluate the performance of our method on the well-known Meta-World benchmark that consists of $50$ complex manipulation tasks. First, we apply MILLION to the ML10 benchmark to compare the performance against state-of-the-art meta-RL algorithms in terms of training and testing success rate. 
Second, we provide an ablation study on ML10 to validate the proposed concepts.
Last, we conduct experiments on the most challenging benchmark ML45 to show its broad effectiveness and generalization capability. 

\begin{figure*}[!t]
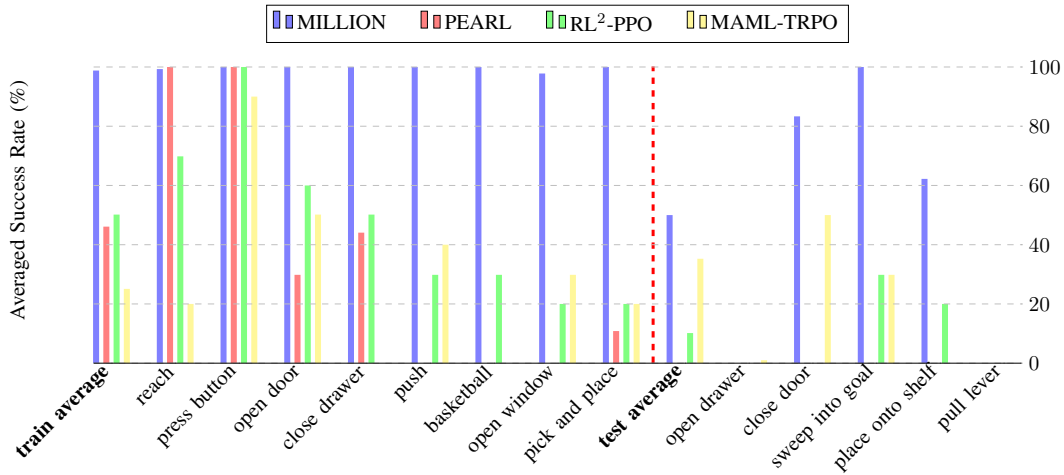

	\drawMLTen
	\caption{Maximum per-task success rates on ML10 V1. MILLION shows the highest performance
		on the training tasks (98.8\%) and the test tasks (50\%). 
	}
	\label{fig:ml10res}
\end{figure*}

\begin{figure*}[!t]
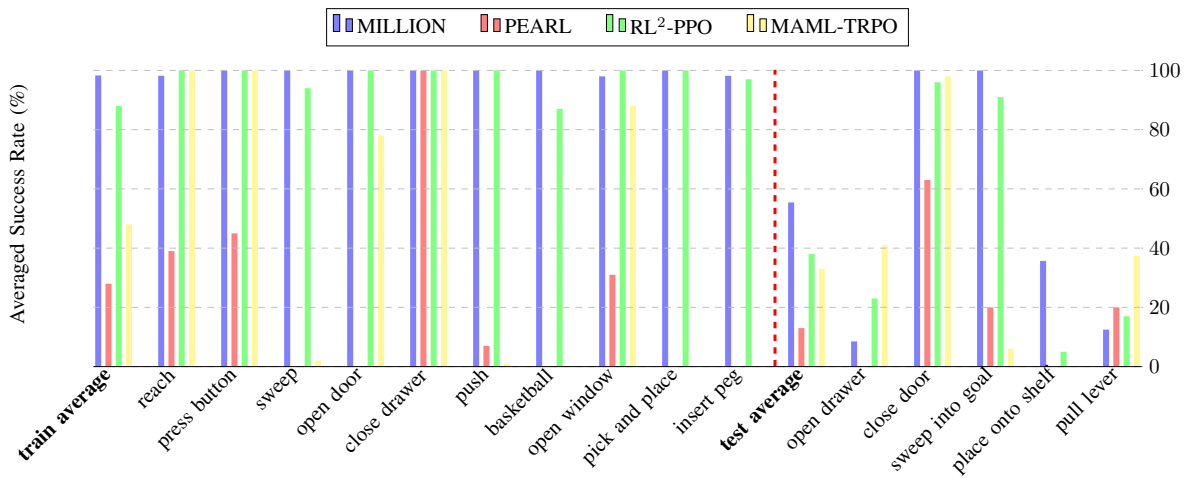

	\drawMLTenV
	\caption{Maximum per-task success rates on ML10 V2. MILLION shows the highest performance
		on the training tasks (98.3\%) and the test tasks (55.4\%). 
	}
	\label{fig:ml10v2res}
\end{figure*}

\subsection{Meta-World Benchmark}

Meta-World \cite{Yu2019a} is a collection of $50$ diverse robotic manipulation tasks built on the MuJoCo physics simulator \cite{Todorov2012}. 
It contains two widely-used benchmarks, namely, ML10 and ML45.
The ML10 contains a subset of the ML45 training tasks, which are split into $10$ training tasks and $5$ test tasks, and the ML45 consists of $45$ training tasks and $5$ test tasks. 
Most tasks contain some kind of object that should be manipulated with the robot arm and adopt the control strategy:
\begin{itemize}
	\item The action space $\mathcal{A}$ contains the desired 3D Cartesian positions of the end-effector and a normalized control command for the gripper.
	\item The state space $\mathcal{S}$ contains the 3D Cartesian positions of the end-effector, the positions of the manipulable objects, and the goal position. 
	The state space is always nine dimensional.
	\item A success metric function is provided for each task, which defines the competition condition of the corresponding task.
	\item For each task, a well-shaped reward function is provided with a similar structure across all tasks, which makes the tasks individually solvable for recent RL algorithms.
\end{itemize}

We make two additional changes to the Meta-World benchmark to reduce the training time. 
First, inspired by \cite{Bellemare2012}, we repeat actions twice during the trial phase to reduce the trial length across all the tasks, which enables a shorter sequence length for the transformer model, and therefore reduces the computation requirements significantly. 
It should be noted that the reported number of environment steps in our results corresponds to the number of observations the agent has seen. 
Second, we add a scalar to the observations during the trial phase, which indicates the remaining time in the trial. 
This helps the agent to learn a better value function, because the Meta-World environments have a time dependent termination condition \cite{Pardo2018}. 
The time observation is computed as $\frac{\text{steps in one trial}}{\text{maximum steps per trial}}$. 
During the instruction phase, a zero value is concatenated to the observation instead. 

There are two versions of Meta-World.
Note that, in the first version of Meta-World, three tasks had to be removed from the benchmark, because the scripted policies provided by Meta-World did not work well to solve the tasks. 
This includes peg-insert-side-v1, lever-pull-v1 and bin-picking-v1. 
Another task, sweep-v1, had to be removed because the reward function did not encourage the agent to solve the task. 
But for the second version of Meta-World, we keep all the tasks accessible for training and testing.

\subsection{ML10 Benchmark \label{ml10_experiments}}

We first tested our method on the ML10 benchmark to show its performance when the agent receives a language instruction instead of only observing the reward signal.
The language instructions are short sentences that describe the goal of the task. 
For each task in ML10, we designed multiple simple language instructions.

\begin{table}[!t]
	\centering
	\caption{Average success rates over all tasks for ML10 and ML45.}
	\begin{tabular}{lllll}
		\toprule
		\multirow{2}{*}{Methods} &
		\multicolumn{2}{c}{ML10} &
		\multicolumn{2}{c}{ML45} \\
		& Training & Testing & Training & Testing \\
		\hline
		MAML & 25\% & 36\% & 21\% & 24\%  \\
		RL$^2$ & 50\% & 10\% & 43\% & 20\%  \\
		PEARL & 43\% & 0\% & 11\% & 30\% \\
		MILLION & \textbf{99\%} & \textbf{50\%} & \textbf{95\%} & \textbf{48\%} \\
		\bottomrule
	\end{tabular}
	\label{tab:ml_results}
\end{table}
\begin{table}[t!]\normalsize
	\centering
	\caption{Comparison among MILLION variants in ML10 V1.}
	\small{
		\begin{tabular}{l|c|c} \hline
			Variants     & Meta-training & Meta-test    \\ \hline
			MILLION    & \textbf{0.99}     & \textbf{0.50}   \\
			without Pop-Art  & 0.41  & 0.30  \\
			without instructions & 0.71 & 0.29 \\
			with Full Time Obs & 0.83 & 0.40 \\ \hline
		\end{tabular}
	}
	\label{tab:ml10_ablation}
	\vspace{-10pt}
\end{table}

\begin{figure}[!t]
	\centering
	\includegraphics[width=0.475\textwidth]{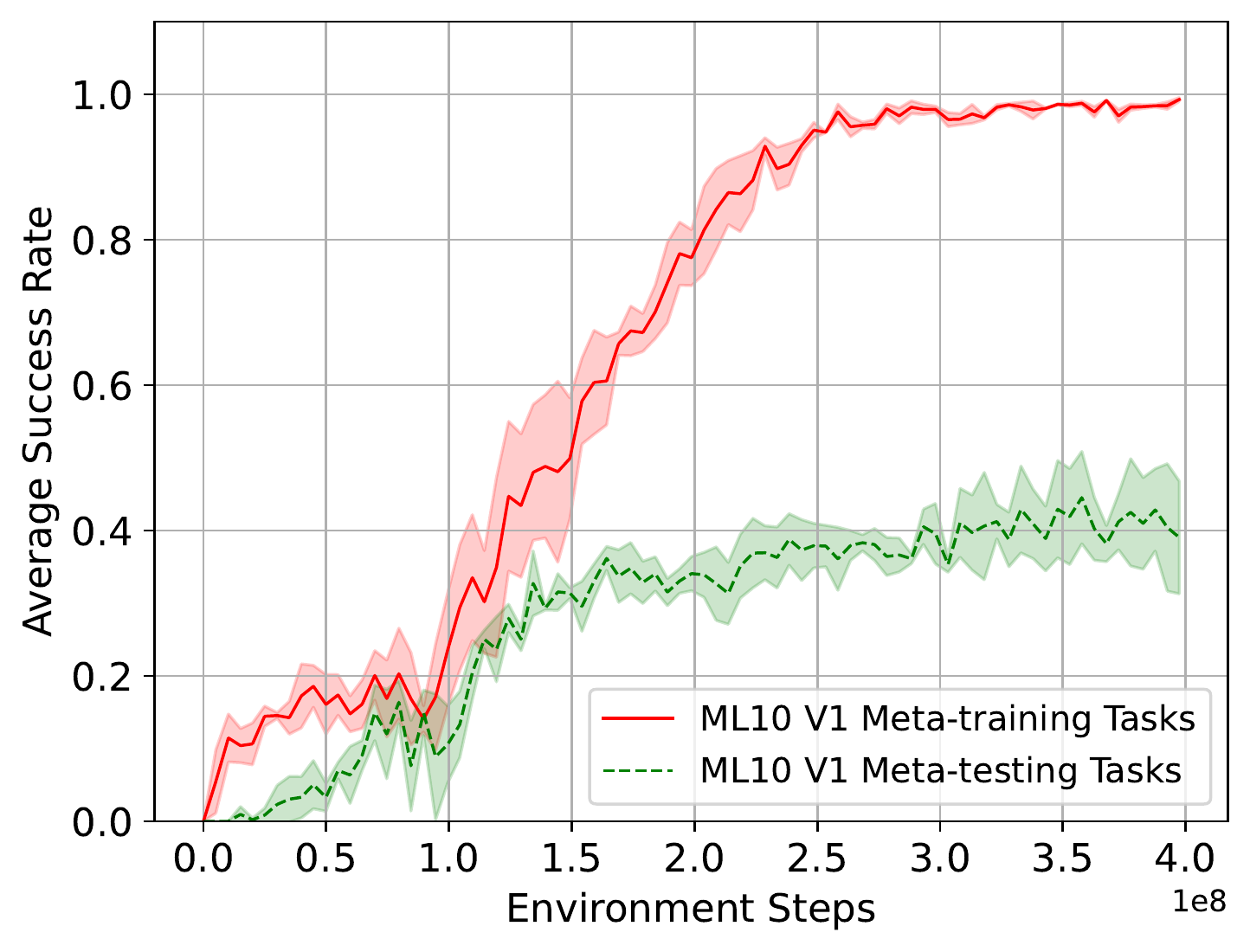}
	\vspace{-0.25cm}
	\caption{Meta-World ML10 V1 training progress with language instructions. The shaded regions indicate one standard deviation over three training runs.  The result of ML10 V2 benchmark can be found on the website\protect\footnotemark[1].}
	\label{fig:metav1}
\end{figure}

According to the reported results from \cite{Yu2019a}, we listed the averaged success rates of state-of-the-art meta-RL algorithms in Table \ref{tab:ml_results}, which includes MAML \cite{finn2017model}, RL$^2$ \cite{duan2017rl}, PEARL \cite{rakelly2019efficient}, and our method MILLION.
Detailed performance for each task in ML10 is visualized in Figure \ref{fig:ml10res} and \ref{fig:ml10v2res}.
It can be observed that, in both versions of Meta-World, MILLION achieves success rates of almost 100\% on the training tasks, which significantly outperforms state-of-the-art methods.
It demonstrates the advantage of providing the agent with the task instructions instead of only rewards.
For meta-testing, MILLION has a success rate of around $50\%$, which also performs better than other methods (See Figure \ref{fig:metav1}.).

\footnotetext[1]{\url{https://tumi6robot.wixsite.com/million}}


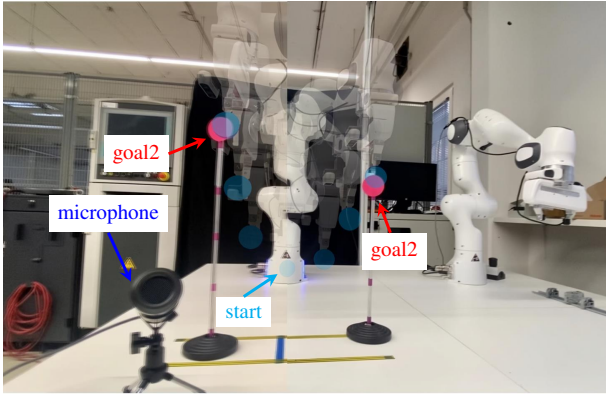
\begin{figure}[!t]
	\centering
	\begin{tikzpicture}
		\centering
		\node[above right, inner sep=0] (image) at (0,0){\includegraphics[width=0.45\textwidth, trim={0.5cm 0 2cm 0},clip]
			{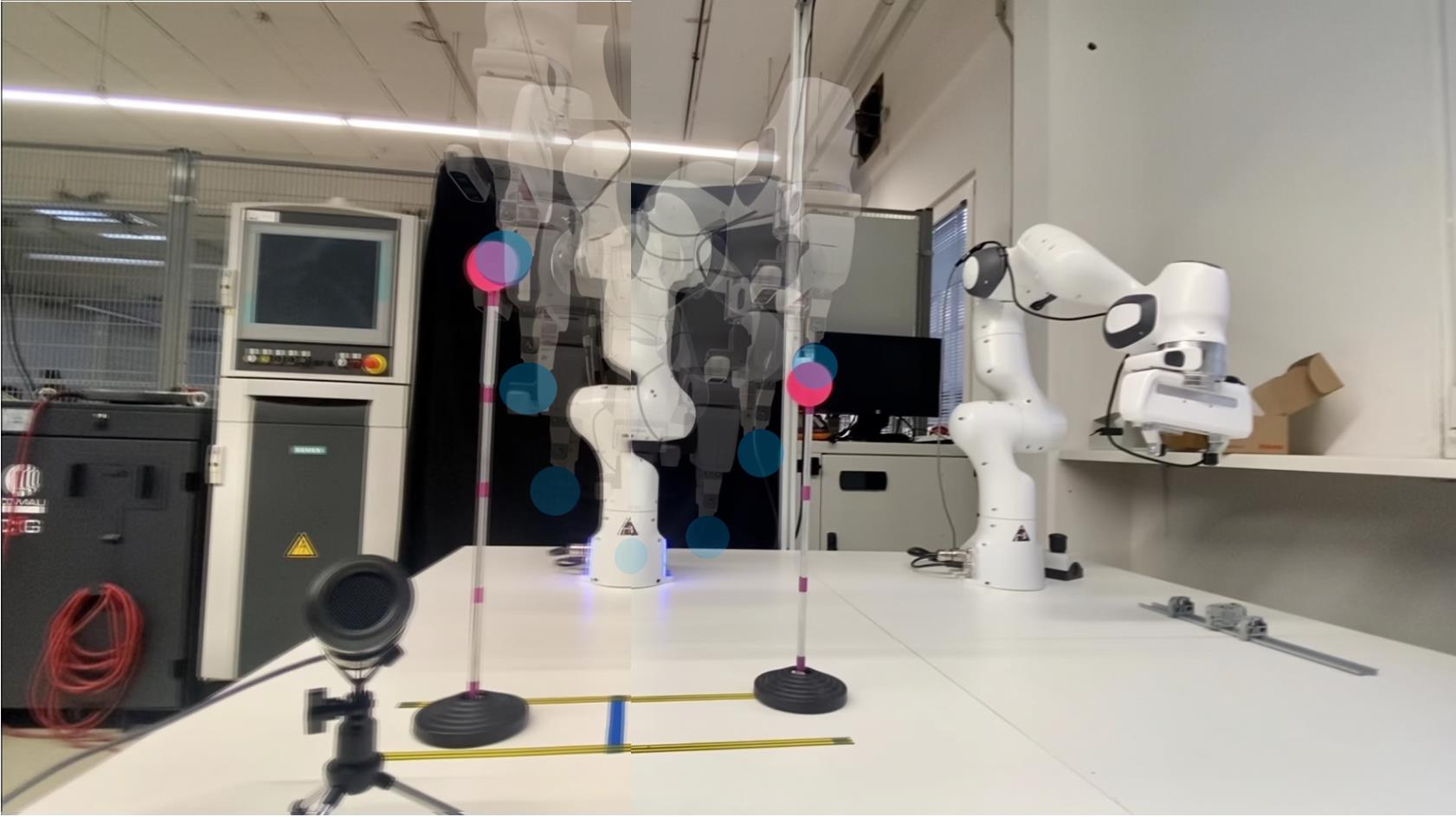}};
		\node(goal1) at (5.2,1.9)[red, fill=white] {\footnotesize {goal2}};
		\draw[arrow1, red, line width=0.4mm] (5.15,2.15) -- (5.0,2.55);
		
		\node(goal2) at (1.77,3.2)[red, fill=white] {\footnotesize {goal2}};
		\draw[arrow1, red, line width=0.4mm] (2.2,3.2) -- (2.7,3.4);
		
		\node(start1) at (3.2,1.1)[cyan, fill=white] {\footnotesize {start}};
		\draw[arrow1, cyan, line width=0.4mm] (3.2,1.3) -- (3.65,1.6);
		\node(micro) at (1.4,2.4)[blue,fill=white!30] {\footnotesize {microphone}};
		\draw[arrow1, blue, line width=0.4mm] (1.4,2.15) -- (1.7,1.5);
	\end{tikzpicture}
	\vspace{-0.25cm}
	\caption{We take the reaching task as one example to show that MILLION can be successfully used in the real world. 
		The task are specified by the language instruction through the microphone.
		More demonstrations can be found on the project webpage\protect\footnotemark[1].}
	\label{fig:realwold_results}
	\vspace{-5pt}
\end{figure}

To emphasize the importance of the proposed concepts, we provide ablation studies on the ML10 benchmark, shown in Table~\ref{tab:ml10_ablation}. 
First, we demonstrate the importance of the Pop-Art reward normalization. 
This normalizes the rewards for every task individually. 
The results demonstrate that Pop-Art is very important for our algorithm. 
Without this, the agent only learns to solve less than half of the training tasks. 
Second, we also examine the performance of a variant that only observes rewards without instructions, which means an episode consists only of three trials and no instruction phase. 
The rewards are simply concatenated to the observations to serve as a potential information source of the task. This is similar to many other recent context-based meta-RL algorithms \cite{duan2017rl, wang2016learning}. 
This variant can learn $70\%$ of the training tasks but adapts to the testing tasks poorly. 
Another ablation is to use a different time observation. Our algorithm observes the remaining time in the current trial. Here we evaluate our algorithm when it observes the remaining time in the full episode. This was originally proposed by \cite{Pardo2018}. The results show that this variant learns the training tasks slightly worse than MILLION. Our hypothesis is that the discount factor causes the value function during a trial to be relatively independent of the next trial rewards. This means that the remaining time is more important for the value function than the remaining time in the episode.

\begin{figure}[!t]
	\centering
	\includegraphics[width=0.475\textwidth]
	{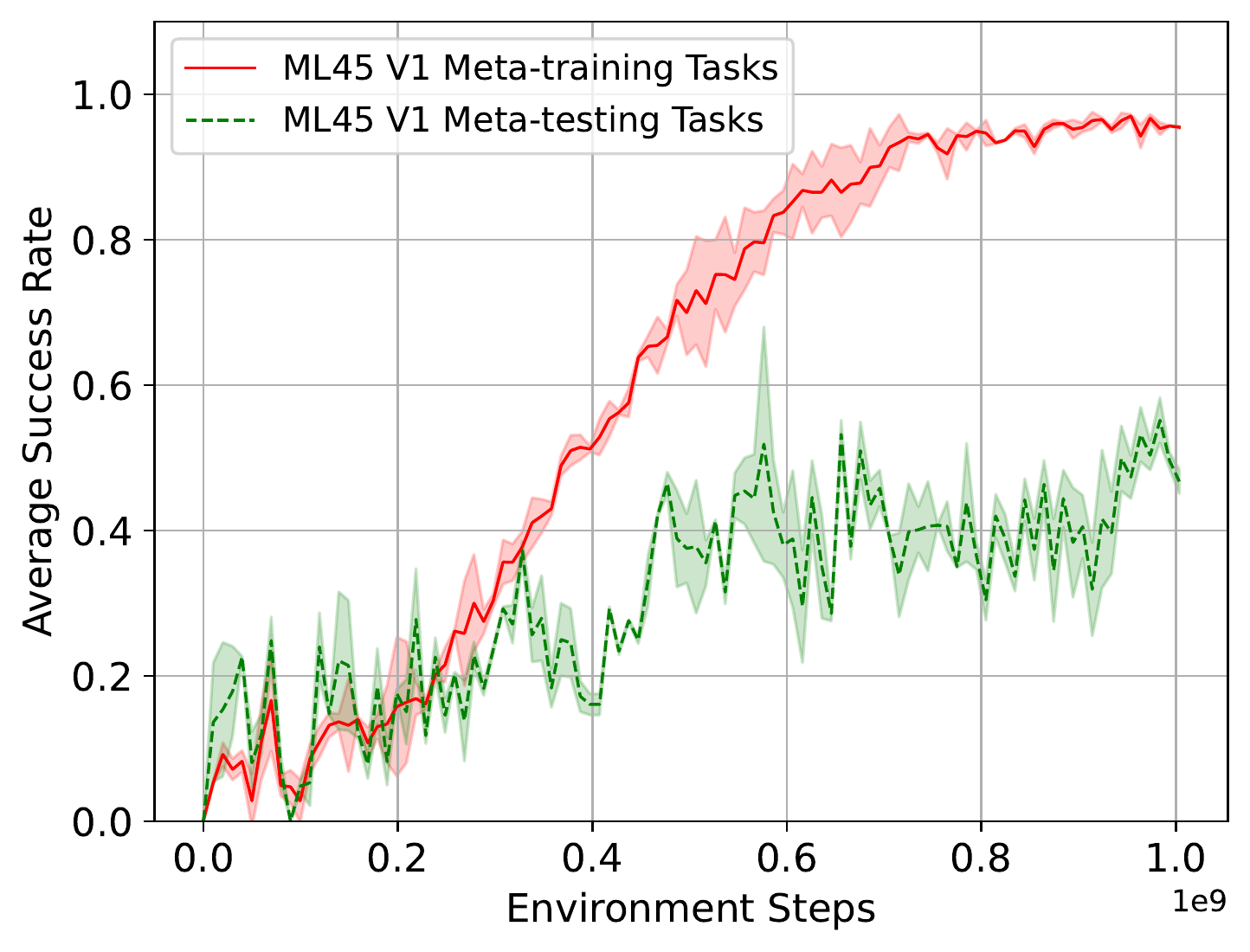}
	\vspace{-0.25cm}
	\caption[Meta-World ML45 training progress]{Average trial success rate on ML45 training and test tasks. The policy is evaluated every 5 million environment steps and averaged over 5 consecutive evaluations.}
	\label{fig:ml45_results}
	\vspace{-5pt}
\end{figure}

We also successfully transfer the learned policy from simulation to the real world.
Due to the page limit, we only show the snapshot of MILLION solving the reach-v1 task in the real world (See Figure \ref{fig:realwold_results}).
More demonstrations of manipulation tasks from ML10 can be found on the webpage\protect\footnotemark[1].

\subsection{ML45 Benchmark}
To test ML 45, we use the same hyperparameters and the same number of trials as for the ML10 benchmark.
However, we train the agent for over 1 billion time steps instead of just 400 million because the benchmark contains more diverse tasks. 
The results (see Figure \ref{fig:ml45_results}) show that MILLION is able to learn almost all training tasks and about 48\% of the test tasks, which indicates that our method has a stable performance on complex manipulation tasks scenarios.
A detailed comparison between MILLION and state-of-the-art algorithms on ML45 is also listed in Table \ref{tab:ml_results}.
It demonstrates that our algorithm MILLION greatly outperforms other baselines in terms of success rate in both training and testing stages.

\section{Conclusion} 
In this paper, we showed that meta-reinforcement learning can be greatly improved by providing the agent with additional task information, such as language instructions, which are often much easier to provide than dense rewards. 
By encoding the language instructions into the observations, we designed a very simple and general algorithm.  
This eases the application of RL algorithms to be used for real-world robotic tasks.
Furthermore, we demonstrated that our algorithm is able to solve a set of very diverse robotic manipulation tasks.
In future, we plan to incorporate the language information as a feedback signal to further calibrate the behavior of the meta-RL agent, which can potentially advance our understanding on interactive intelligent robots in the future.

\bibliographystyle{IEEEtran}
\bibliography{icra}

\end{document}